# Enhancing Machine Translation through Advanced In-Context Learning: A Methodological Strategy for GPT-4 Improvement

Yufeng Chen

School of International Liberal Studies, Waseda University, Tokyo, 169-8050, Japan

www1500977658@toki.waseda.jp

**Abstract.** The challenge of improving translation accuracy in GPT-4 is being addressed by harnessing a method known as in-context learning. This paper introduces a strategic approach to utilize in-context learning specifically for machine translation, aiming to significantly boost accuracy. The crux of this method lies in the judicious selection of demonstrations that are most effective for in-context learning. By selecting these examples carefully, GPT-4 can utilize them to achieve remarkably accurate machine translations, eliminating the need for task-specific fine-tuning. This technique is anchored in the semantic similarities between the user's prompt and the chosen dataset. Sentences from this dataset, carefully picked for their relevance and clarity, serve as potent demonstrations for in-context learning. This approach not only enhances translation accuracy but also enriches the understanding of nuanced linguistic structures. It represents a significant step forward in machine learning, leveraging the inherent capabilities of GPT-4 to provide translations that are not only accurate but also contextually rich and linguistically sophisticated. This method demonstrates the potential of in-context learning in overcoming language barriers, opening new avenues for cross-cultural communication and global collaboration.

## 1. Introduction

Large language models (LLMs) like GPT-4 have shown remarkable skills in various tasks, notably through in-context learning, a method characterized by conditioning on a limited set of input-label pairs (Brown et al. 2020) [1]. This approach allows GPT-4 to enhance its performance in specific tasks such as translation without the need for additional fine-tuning. Essentially, GPT-4's ability to understand and execute tasks surpasses that of earlier models not utilizing in-context learning. This superior performance stems from the model's advanced algorithms and architecture, which enable it to assimilate and apply new information efficiently. By feeding GPT-4 with relevant examples pertaining to a particular task, it can adapt and improve its output, demonstrating a sophisticated understanding of context and nuances. This adaptability is crucial in handling complex, varied tasks, making GPT-4 a versatile tool in natural language processing. Its capability to learn and adapt in this manner showcases the evolution of artificial intelligence, where models can intuitively enhance their abilities through exposure to specific examples, aligning closely with how human learning occurs. As shown in Fig 1.

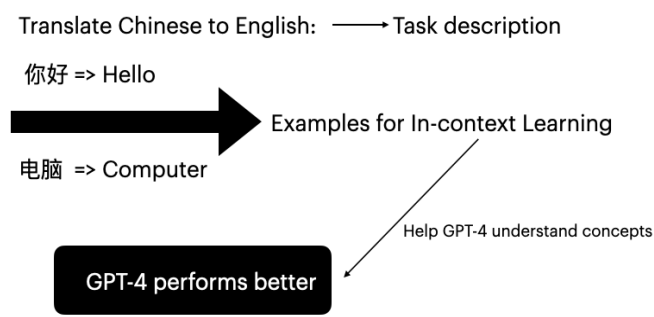

**Fig. 1** To achieve in-context learning for Chinese to English translation task, a few-shot approach involving specific task examples is employed (Photo/Picture credit: Original).

In fact, the potency of in-context learning as a useful tool for LLMs stems from the equation grounded in Implicit Bayesian Inference.

$$P(\text{output}|\text{prompt}) = \int_{\text{concept}} P(\text{output}|\text{concept}, \text{prompt}) P(\text{concept}|\text{prompt}) d(\text{concept}) \tag{1}$$

According to this equation, the output of LLMs could be better by selecting the prompt concept more effectively [2, 3]. Without a doubt, the random selection of examples cannot effectively facilitate GPT-4 in acquiring a comprehensive understanding of the prompt concept [4-6]. Consequently, the primary objective becomes the strategic selection of more suitable examples based on the user's input prompt, thereby enhancing GPT-4's performance. In the subsequent section, this paper will present a methodology designed to facilitate the selection of improved translation examples from a dataset based on the input. This approach aims to empower GPT-4 to achieve high-accuracy translations from Chinese to English (ZH to EN), Japanese to English (JA to EN), and Vietnamese to English (VI to EN).

## 2. Proposed Method

This methodology utilizes a dataset, referred to as Dselect, comprising language translation pairs, specifically selected for in-context learning. The dataset, potentially vast in size, plays a critical role, as this research delves into the impact of its magnitude on translation outcomes. Central to this approach is the design of a text retriever, tasked with identifying and extracting the top K sentences from Dselect. These sentences should closely align in meaning with the user's input. As per reference [7], the retrieval process involves two primary components: a TF-IDF matrix and the application of cosine similarity measures. A detailed exploration of these components will be provided. The top K sentences, as determined by the retriever, are then amalgamated with the user's input. Following this, the GPT-4 model steps in to perform the translation. To evaluate the translation's precision, metrics like BLEU and COMET are utilized. This approach underscores the intricate interplay between dataset size, retrieval mechanisms, and translation accuracy in machine learning frameworks.

### 2.1 TF-IDF Score

The TF-IDF matrix is composed of TF-IDF scores. And TF-IDF scores can be calculated as $TF(t, d) = \frac{Number\ of\ times\ term\ appears\ in\ document\ d}{Total\ number\ of\ terms\ in\ document\ d}$, which represents the term frequency that measures how often a word appears in a document. And, the IDF also needs to be considered. $IDF(t,D) = \log(\frac{Total\ number\ of\ documents\ in\ the\ corpus\ D}{Number\ of\ documents\ containing\ term\ t})$, which measures the significance of a word across a collection of documents. In the present study, the symbol "D" is employed as the selected dataset, denoted as Dselect, where "d" signifies an individual sentence within the confines of Dselect. Based on these pieces of information, the TF-IDF scores can be calculated to construct the TF-IDF matrix eventually. Specifically, the TF-IDF scores are determined through the expression TF (t, d)×IDF (t, D), allowing for the quantification of the significance of a given word within a particular document.

**2.1.1 Cosine Similarity**

A technique used to evaluate the resemblance between two vectors in an inner product space, finds significant application in the realm of TF-IDF vectors. Particularly in the evaluation of document similarity, it gauges the similarity between documents by considering the angle between their vector representations. The cosine similarity between vectors A and B is computed using the formula: cosine similarity (A, B) = $\dfrac{A \cdot B}{\|A\| \cdot \|B\|}$ . In this study, "A" corresponds to the TF-IDF vector of the user prompt while B corresponds to the TF-IDF vectors of other documents within the dataset Dselect. It is evident that a higher cosine similarity score signifies a greater likeness between the user prompt and other documents. As a result, the top K examples can be selected from the dataset Dselect based on their similarity scores to serve as in-context learning examples. As shown in Fig 2.

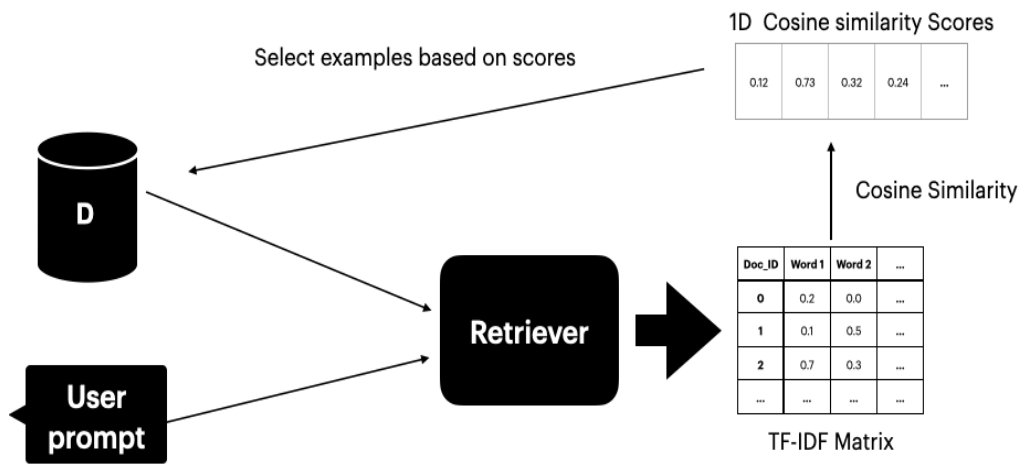

**Fig. 2** Use retriever to create the TF-IDF matrix and cosine similarity scores to select the top-K examples from Dselect for in-context learning (Photo/Picture credit: Original).

## 3. Experimental Setup

### 3.1 Experimental procedure

The methodology under discussion aims to assess GPT-4's translation capabilities across three language pairs: Chinese to English (ZH-EN), Japanese to English (JA-EN), and Vietnamese to English (VI-EN). This assessment is structured to unfold in three distinct experimental scenarios. In the initial scenario, the translation process is executed without integrating in-context learning. This approach means no prior examples are provided, challenging GPT-4 to translate based solely on its pre-existing knowledge and algorithms. This baseline scenario establishes a fundamental understanding of GPT-4's inherent translation capabilities.

The second scenario introduces a twist to the in-context learning process. Here, examples are incorporated into the translation task. However, these examples are selected randomly. This randomness means that the relevance of these examples to the current translation task is left to chance. This scenario tests the adaptability of GPT-4 in utilizing random contextual clues to enhance translation accuracy. The third and most innovative scenario implements a specialized method. A retriever computes a Term Frequency-Inverse Document Frequency (TF-IDF) matrix. This matrix is instrumental in calculating cosine similarity scores between the user prompt and various sentences in the dataset. Based on these scores, the top four most relevant examples are identified and fed to GPT-4 as context for the translation task. This targeted approach to in-context learning is hypothesized to significantly refine the translation output. Upon completion of the translations in each scenario, two advanced evaluation metrics will be employed to measure GPT-4's performance: the Bilingual Evaluation Understudy (BLEU) and the Cross-lingual Optimized Metric for Evaluation of Translation (COMET). BLEU focuses on the precision of word choice and phrase matching, while COMET provides a more holistic assessment, considering factors like fluency and semantic accuracy. The application of both metrics ensures a comprehensive evaluation, capturing various facets of translation quality. This dual-metric approach will offer an in-depth understanding of how GPT-4 fares in translating across the selected language pairs, under varying degrees of contextual support. The results are anticipated to reveal valuable insights into the effectiveness of the proposed method and its impact on enhancing machine translation capabilities.

### 3.1.1 BLEU Score

The BLEU Score, a bilingual evaluation understudy, is determined for each translated segment by comparing it to reference translations. These scores are averaged across the entire corpus to evaluate the overall translation quality [8]. Remarkably, this method aligns with human quality judgments, making BLEU a reliable metric for assessing GPT-4's translation accuracy.

### 3.1.2 COMET Score

The COMET score, a neural framework, is designed for multilingual machine translation evaluation, achieving high correlation with human assessments [9]. It requires three inputs: the translated text, the original text, and the reference translation. These are encoded by a pre-trained encoder and processed through a feed-forward regressor for evaluation..

### 3.2 Datasets

OPUS-100 was selected for its comprehensive range of translation language pairs (e.g., ZH-EN, JA-EN, VI-EN) and diverse domains, fulfilling the study's requirements without needing multiple datasets [10]. The OPUS-100 dataset was divided into two segments: 10,000 training instances for each language pair and the first 100 sentences from the testing dataset of OPUS-100 for each pair were used for testing.

### 3.3.Programming Code

The programming code involves simple steps. Initially, TfidfVectorizer and cosine_similarity functions are imported from the scikit-learn package. These functions are used to create the retriever. In the retriever, a key step is combining the user prompt with Dselect to calculate cosine similarity scores between the prompt and all sentences in Dselect. This enables identifying the top 4 examples from Dselect based on the prompt, which are then embedded into GPT-4.

```
messages = [
    {"role": "system", "content": "You are a translation assistant from Chinese to English. Some rules to remember:\n\n- Do not add extra blank lines.\n- It is important to maintain the accuracy of the contents, but we don't want the output to read like it's been translated. So instead of translating word by word, prioritize naturalness and ease of communication.\n\n Here are some examples that you can use to learn how to translate from Chinese to English:\n"+ example_translations_str},
    {"role": "user", "content":f' Please translate the given Chinese sentence {text} to English sentence and please make the translation as accurate and natural as possible.' }
]
```

**Fig. 3** The ultimate prompt instructs GPT-4 to perform Chinese-to-English translation, incorporating the optimal four examples from Dselect (Photo/Picture credit: Original).

In Fig 3, the prompt, enriched with in-context learning, is now complete, featuring the four examples identified by the retriever. Consequently, the subsequent phase involves the analysis of results derived from the experiment.

## 4. Results and Discussion

Table 1 summarizes all results. Based on these findings, the approach demonstrates superior translation accuracy compared to other scenarios across all three language pairs. Despite the seemingly modest increase, a 1% improvement in BLEU score holds significant importance in the context of machine translation. It is noteworthy that the effectiveness of random in-context learning occasionally lags behind the scenario of not employing in-context learning at all. This highlights the critical importance of judiciously selecting examples for GPT-4 during the in-context learning process, as inappropriate examples may adversely affect its overall performance.

**Table 1.** Illustrates the translation accuracy outcomes across all three distinct scenarios for all language pairs.

| Evaluation Matrix | COMET | BLEU |
|---|---|---|
| ZH-EN | | |
| Without ICL | 0.8081 | 0.2515 |
| Random ICL | 0.8078 | 0.2687 |
| Retrieve ICL | 0.8195 | 0.2922 |
| JA-EN | | |
| Without ICL | 0.7184 | 0.2163 |
| Random ICL | 0.7140 | 0.1909 |
| Retrieve ICL | 0.7395 | 0.2374 |
| VI-EN | | |
| Without ICL | 0.7316 | 0.2451 |
| Random ICL | 0.7332 | 0.2707 |
| Retrieve ICL | 0.7511 | 0.2877 |

And another aspect to consider is the size of Dselect. At the beginning, the expectation was that a larger Dselect would yield better results, as a sizable dataset has the potential to encompass a diverse range of domains, providing more effective examples for GPT-4. This assumption was validated through experimentation with a larger Dselect comprising 1 million sentences. Table 2 illustrates the results obtained when using this expanded dataset as Dselect for selecting in-context learning examples.

**Table 2.** Illustrates the variations in translation accuracy corresponding to the incremental augmentation of the Dselect dataset size.

| Evaluation Matrix | COMET | BLEU |
|---|---|---|
| ZH-EN | | |
| Retrieve ICL Dselect=10000 | 0.8195 | 0.2922 |
| Retrieve ICL Dselect=1000000 | 0.8313 | 0.3016 |

There is no doubt that using a larger dataset as Dselect holds the potential to enhance the efficacy of task learning for GPT-4. Therefore, the amalgamation of this methodology with a crafted extensive dataset becomes imperative for enabling GPT-4 to attain high performance, particularly in the domain of machine translation.

## 5. Conclusion and Next Steps

This paper introduces an innovative method to enhance GPT-4's translation capabilities through in-context learning. The core of this approach is building a retriever using a TF-IDF matrix and cosine similarity scores. This retriever identifies sentences in the Dselect dataset closely matching the user prompt. Selected examples from Dselect then support GPT-4's in-context learning. Experimental evaluations show this method's effectiveness, with notable improvements in BLEU and COMET scores compared to scenarios lacking in-context learning or using random examples. Furthermore, a larger Dselect dataset significantly boosts this method's efficiency by providing a wider range of potential examples. However, two critical areas require further investigation. The first is developing a robust dataset, referred to as Dselect. Although OPUS-100 was used in this study, creating a comprehensive dataset with diverse domains and accurate translation references is crucial. This time-intensive process is expected to significantly improve GPT-4's translation proficiency. The second area of exploration is the impact of the number of in-context learning examples on translation accuracy. Currently, the method utilizes the top 4 examples based on cosine similarity scores. Future research examining the effects of using 5 or 10 examples will shed light on how example quantity influences accuracy. This paper, focusing on the quality of in-context learning examples, sets the stage for further research and practical application developments.